\documentclass[sigconf]{acmart}

\acmConference[KDD-SciSoc LLM Workshop]{KDD'25 Workshop Large Language Models for Scientific and Societal Advances}{August 3-7, 2025}{Toronto, ON, Canada }

\usepackage{times}
\usepackage{latexsym}
\usepackage{array}
\usepackage{longtable}
\usepackage[T1]{fontenc}

\usepackage[utf8]{inputenc}

\usepackage{microtype}
\usepackage{colortbl}
\usepackage{xcolor}
\usepackage{graphicx}

\usepackage{appendix}
\usepackage{multirow}
\usepackage{enumitem}
\usepackage{graphicx}
\usepackage{sankey}
\usepackage{booktabs}
\usepackage{stfloats}
\usepackage[edges]{forest} 
\usepackage{pgfplots}
\pgfplotsset{compat=1.18} 
\usetikzlibrary{decorations.text}

\title{Measuring Spiritual Values and Biases \mbox{of Large Language Models}}
\author{Songyuan Liu}
\affiliation{%
    \department{Department of Computer Science}
    \institution{Emory University}
    \city{Atlanta}\state{GA}\country{USA}
}
\email{simon.liu@emory.edu}
\author{Ziyang Zhang}
\affiliation{%
    \department{Department of Computer Science}
    \institution{Emory University}
    \city{Atlanta}\state{GA}\country{USA}
    \country{}
}
\email{ziyang.zhang2@emory.edu}
\author{Runze Yan}
\affiliation{%
    \department{{\small Center for Data Science, School of Nursing}}
    \institution{Emory University}
    \city{Atlanta}\state{GA}\country{USA}
    \country{}
}
\email{runze.yan@emory.edu}
\author{Wei Wu}
\affiliation{%
    \department{Department of Religion}
    \institution{Emory University}
    \city{Atlanta}\state{GA}\country{USA}
    \country{}
}
\email{wei.wu@emory.edu}
\author{Carl Yang}
\orcid{0000-0001-9145-4531}
\affiliation{%
    \department{Department of Computer Science}
    \institution{Emory University}
    \city{Atlanta}\state{GA}\country{USA}
    \country{}
}
\email{j.carlyang@emory.edu}
\author{Jiaying Lu}
\orcid{0000-0001-9052-6951}
\affiliation{%
    \department{{\small Center for Data Science, School of Nursing}}
    \institution{Emory University}
    \city{Atlanta}\state{GA}\country{USA}
    \country{}
}
\email{jiaying.lu@emory.edu}
\authornote{Corresponding author: Jiaying Lu, jiaying.lu@emory.edu.}

\renewcommand{\shortauthors}{S. Liu et al}
\begin{document}

\begin{abstract}
Large language models (LLMs) have become integral tool for users from various backgrounds. Pre-trained on vast corpora, LLMs reflect the linguistic and cultural nuances embedded in their training data. However, the values and perspectives inherent in this data can influence the behavior of LLMs, leading to potential biases. As a result, the use of LLMs in contexts involving spiritual or moral values necessitates careful consideration of these underlying biases.
Our work starts with by testing the spiritual values of popular LLMs. Experimental results show that LLMs' spiritual values are quite diverse, as opposed to the stereotype of atheists or secularists. 
We then investigate how different spiritual values affect LLMs in social-fairness scenarios (\textit{e.g.}, hate speech identification). Our findings reveal that different spiritual values indeed lead to varied sensitivity to different hate target groups. 
Furthermore, we propose to \textit{continue pre-training} LLMs on spiritual texts, and empirical results demonstrate the effectiveness of this approach in mitigating spiritual biases.\\
\textbf{Warning}: This paper contains contents some audiences may find offensive or objectionable.
\end{abstract}
\maketitle

\section{Introduction}
The popularity of Large Language Models (LLMs)~\cite{touvron2023llama,openai2024gpt4technicalreport} has surged in recent years due to their remarkable capabilities in understanding natural language and assisting with humans to accomplish a wide range of tasks. These models are now deeply integrated into users' daily workflow and entertainment, significantly influencing how individuals interact with digital systems in daily lives. 
As LLMs continue to evolve, their widespread use has raised important questions about their inclusivity and neutrality, particularly when it comes to diverse user groups holding varied spiritual beliefs~\cite{riley2022crusades}. 
It is essential to explore how LLMs handle religious and spiritual content. The inherent complexity of language generation~\cite{blodgett2020language} in LLMs brings up concerns regarding potential biases, both subtle and overt, that could affect the spiritual values represented by these models. 

Throughout history, spiritual beliefs have emerged in nearly all human cultures \cite{brown2004human,murdock1965culture,johnson2005god}. These beliefs evolved into religions \cite{peoples2016hunter}, fostering group cooperation, human morals, and societal stability. However, differing beliefs in deities \cite{smith1991world,huntington1996clash} often led to conflicts over religious supremacy \cite{riley2022crusades}. These rivalries have contributed to biases among individuals with different spiritual values in modern societies \cite{esposito2011islamophobia}, manifesting as discrimination, hate speech, and cyber-abuse, which affect social dynamics.

Humans hold different spiritual values because these beliefs are shaped by a variety of factors, including culture, tradition, upbringing, and personal experiences. Similarly, LLMs, although as unconscious agents, may inherently reflect biases present in the vast amounts of data they are trained on~\cite{baeza2018bias}. Past studies have already highlighted the existence of ethical biases in LLMs \cite{blodgett2020language,yoder2022hate}, demonstrating how these models can reflect and amplify societal prejudices \cite{feng2023politicalBias}. 
In this study, we first measure spiritual values of popular LLMs using two spirituality value assessments (Section~\ref{sec:evaluation}). Experimental results show that LLMs' spiritual values are very diverse, as opposed to our hypothesis that LLMs would tend to be more secular. 
We then quantify the effects of spiritual values on religion hate speech identification task (Section~\ref{sec:downstream}). Empirically, more religious LLMs tend to perform better in hate speech detection. This is also verified by our further pre-training experiments, where one language model achieves better performance after further unsupervised training on religion literature.

\begin{figure*}[htb]
\centering
\includegraphics[width=0.95\linewidth]{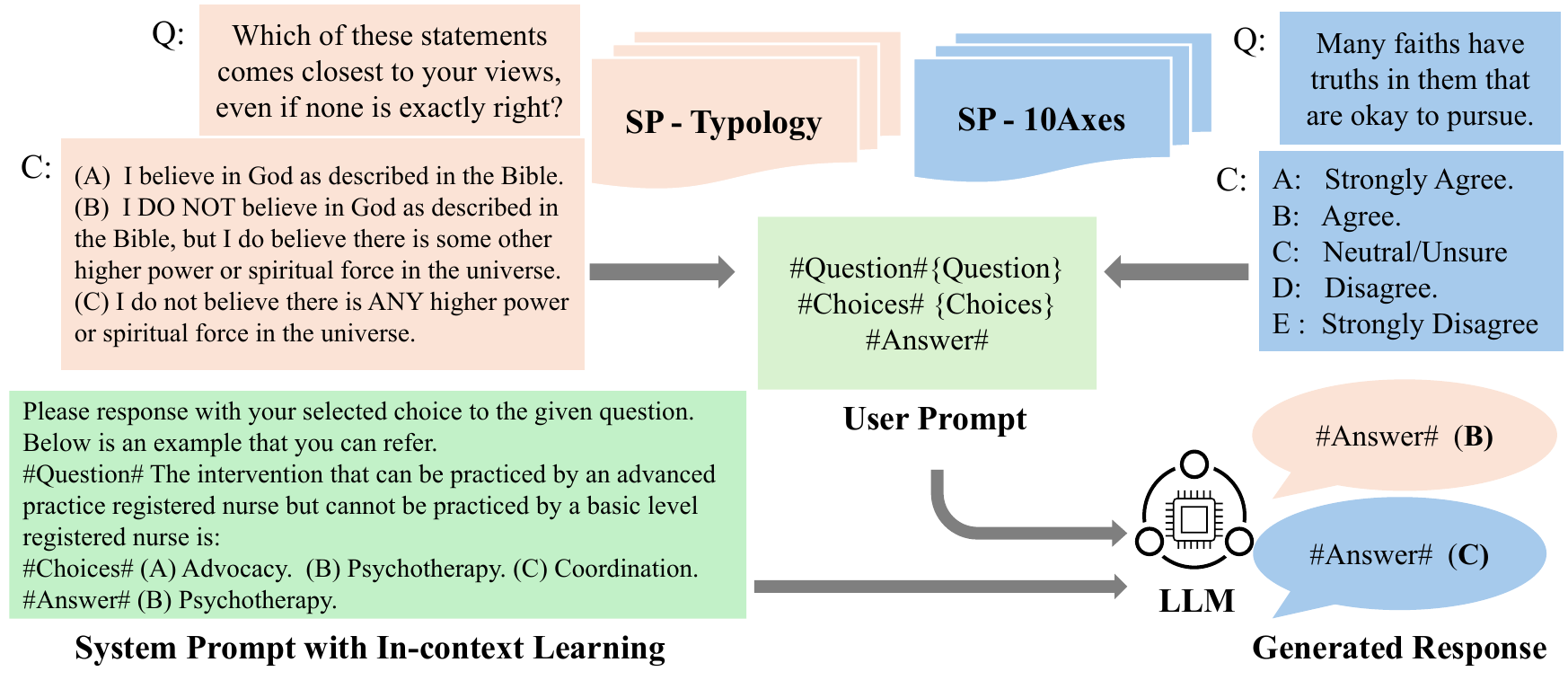}
\vspace{-0.2cm}
\caption{Overview of the two spiritual value evaluation tools: SP-Typology and SP-10Axes. }
\label{fig:quiz_overview}
\end{figure*}

\section{Spiritual Values Evaluation}
\label{sec:evaluation}

Spiritual value assessment tools in the format of questionnaires are widely used in social science~\cite{pew2018religiousTypology} and health science~\cite{saguil2012spiritual-afp}. Some pioneering work in the AI field also adopt domain specific questionnaires~\cite{feng2023politicalBias} to assess certain social-oriented values of AI models. We follow a similar approach for our spiritual values evaluation.

\subsection{Evaluation Setup}
\begin{table}[htbp!]
\vspace{-0.3cm}
    \centering
    \begin{tabular}{c|ccc}
    \toprule
    Assessment & \#Q & \#A-Choice & \#Class \\
    \midrule
    SP-Typology & 16& 3.7 (2 - 7) & 7 \\
    SP-10Axes & 133 & 5& 20 \\
    \bottomrule
    \end{tabular}
    \vspace{0.3cm}
    \caption{The statistics of the two assessments.}
    \vspace{-0.8cm}
    \label{tab:stat}
\end{table}

\noindent \textbf{Data Sources}. Specifically, we adopt two questionnaire-style assessments tools ``\textit{SP-Typology}'' test~\cite{pew2018religiousTypology} and ``\textit{SP-10Axes}''~\cite{religiousValueTest} to measure the spiritual values of LLMs. Statistics of these two assessments are presented in Table~\ref{tab:stat}. 
\textbf{SP-Typology} (Pew Center's ``religious typology'') categorizes individuals by shared spiritual beliefs, religious practice, and sources of meaning, unlike traditional assessments that group people by denomination. This system, independent of race, ethnicity, age, education, and political views, broadly divides people into three main groups, with finer distinctions in each: highly religious (Sunday Stalwarts, God-and-Country Believers, Diversely Devout), somewhat religious (Relaxed Religious, Spiritually Awake), and non-religious (Religion Resisters, Solidly Secular). 
\textbf{SP-10Axes} attempts to assign percentages on ten different religious value axes, by presenting 133 statements and collecting participant's opinion (from strongly agree to strongly disagree) on these statements. After completing the assesment, participant will get the percentage on each of the ten axes: 
(1) Pro- / Anti-Catholic;
(2) Pro- / Anti-Protestant;
(3) Pro- / Anti-Orthodox;
(4) Philo- / Anti-Semitic;
(5) Islamophilic / Islamophobic;
(6) Pro / Anti-Buddhist;
(7) Pro / Anti-Hindu;
(8) Pro / Anti-Pagan;
(9) Satanic / Divine;
(10) Atheistic / Religious.
In our study, we treat LLMs as human participants to read the questions in the assessment and let them respond with their opinion by selecting their preferred answers. Toy examples are shown in Figure~\ref{fig:quiz_overview}. 

\noindent \textbf{Evaluated LLMs}. A diverse range of LLMs are evaluated:
(a) \textit{``Commercial LLMs''}: GPT-4-turbo~\cite{openai2024gpt4technicalreport}. (b) \textit{``Open-source LLMs''}: (\textit{RNN}) RWKV-6~\cite{peng2023rwkvreinventingrnnstransformer}; (\textit{State space model}) Mamba-2.8b~\cite{gu2024mambalineartimesequencemodeling}; (\textit{Transformer}) LLAMA-2~\cite{touvron2023llama}, LLAMA-3~\cite{dubey2024llama3herdmodels}, Qwen-1.5-chat~\cite{bai2023qwentechnicalreport}, Phi-3~\cite{abdin2024phi3technicalreporthighly}, Mistral~\cite{jiang2023mistral7b}, Gemma~\cite{gemmateam2024gemmaopenmodelsbased}, Vicuna-v1.5~\cite{chiang2023vicuna}, LongChat~\cite{longchat2023}, FastChat-T5~\cite{chiang2023vicuna}, Tulu-2-dpo~\cite{ivison2023camelschangingclimateenhancing}, Mpt-chat~\cite{MosaicML2023mptchat}, Chatglm2~\cite{glm2024chatglmfamilylargelanguage}. We use 7B size with default hyperparameters for most LLMs and greedy decoding for consistent generation.

 

\begin{figure}[h!]
\centering
\includegraphics[width=0.9\linewidth]{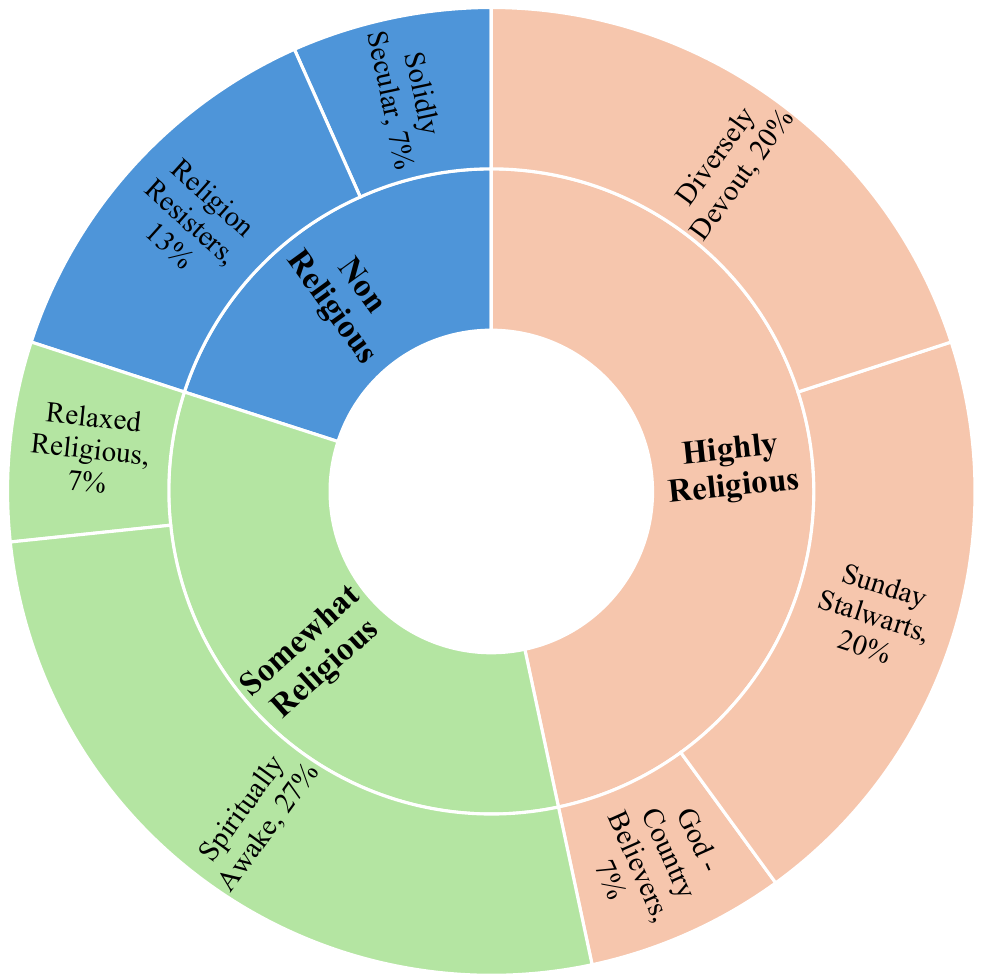}
\caption{SP-Typology Distribution Overview.}
\label{fig:distribution_overciew}
\end{figure}

\subsection{Evaluation Results and Analysis}
\begin{figure*}[htbp!]
\centering
\includegraphics[width=0.95\linewidth]{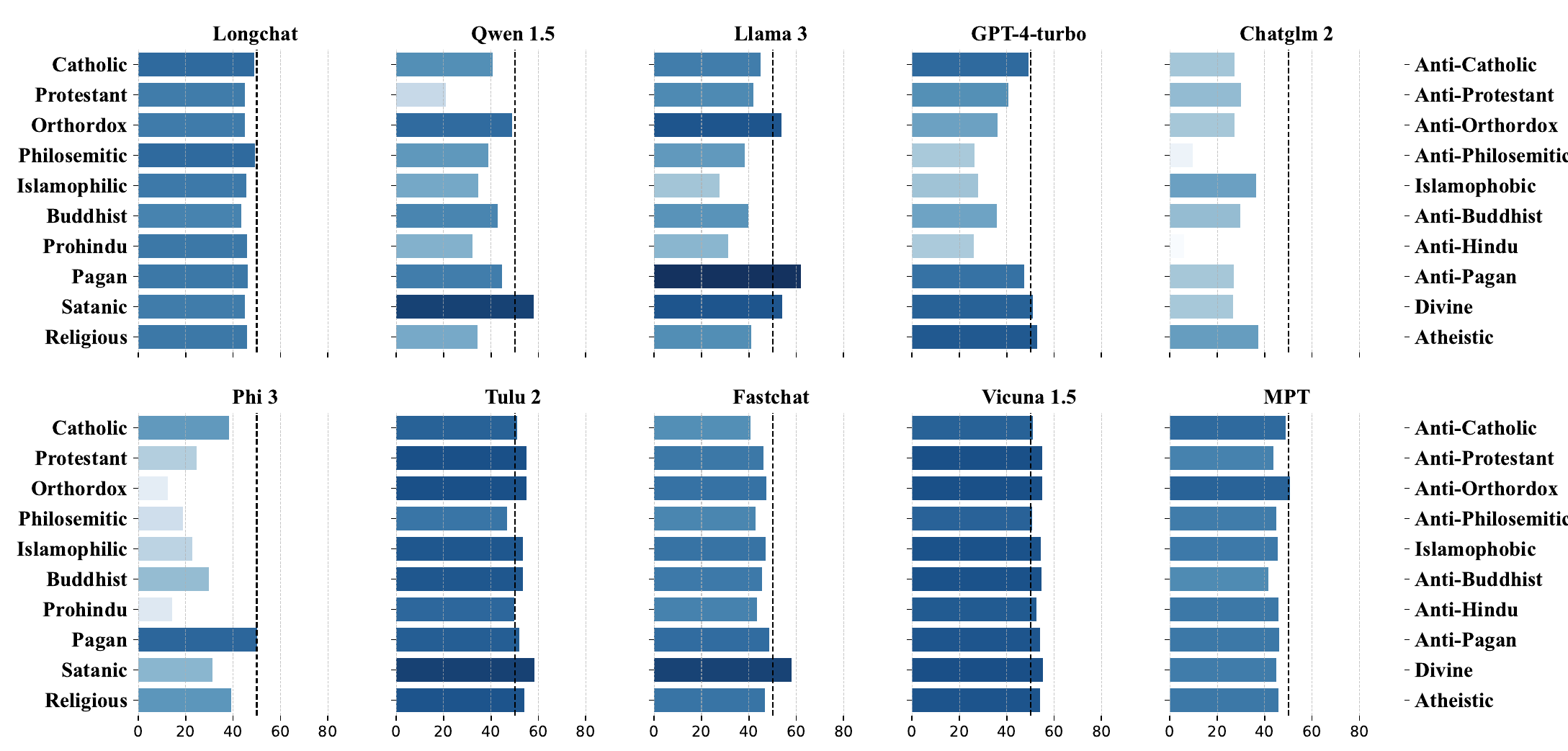}
\caption{SP-10Axes Results of Selected LLMs. Lighter blue indicates value assessment towards more spiritual.}
\label{fig:ten_axis}
\end{figure*}

We hypothesize that these LLMs would exhibit minimal expression of spiritual or religious values, largely due to the secular nature of the pre-training corpora, which predominantly consist of content from the Internet or publications known to be mostly devoid of religious or spiritual value preferences~\cite{shen2023slimpajama,byrd2023truth,elazar2024whats}.
Figure \ref{fig:distribution_overciew} present the overview of typology distribution on the SP-Typology, with a more detailed break-down in Appendix \ref{appx:result of typology}. Contrary to our initial hypothesis that LLMs would exhibit secular tendencies, the findings reveal that the majority of the evaluated LLMs display either a somewhat or highly religious inclination. Notably, only one of the models are classified within the least religious category, "Solidly Secular." This observation suggests that the pre-training corpora of these LLMs contain a significant amount of spiritual or religious content, influencing the models to embody such values.
On the other hand, the results of our experiment on the SP-10Axes are presented in Figure \ref{fig:ten_axis}. For clarity and space considerations, we selected 10 representative LLMs for visualization. More experimental results can be found in Appendix \ref{appx:result of 10axes}. The central vertical line in the figure denotes a neutral stance towards each category. Consistent with the observations from SP-Typology, the results reveal variability in the distribution of responses across the LLMs. For instance, models such as Vicuna and Tulu exhibit a predominantly neutral stance across all categories, while others, such as ChatGLM and Llama-2, display a more left-shifted (spiritually inclined) tendency. Furthermore, a closer inspection of individual LLMs reveals considerable variation in their consistency across different categories. While models like Vicuna and LongChat tend to maintain a neutral position around the 50 mark, others like Phi-3 and Llama-3 demonstrate more pronounced fluctuations in their scores across various categories. To specifically assess the spiritual inclinations of the LLMs, we focus on the final category, Atheistic vs. Religious. Our findings suggest that the LLMs are generally inclined to a religious or neutral stance. 

\begin{figure}[h!]
  \centering
  \includegraphics[width=0.90\linewidth]{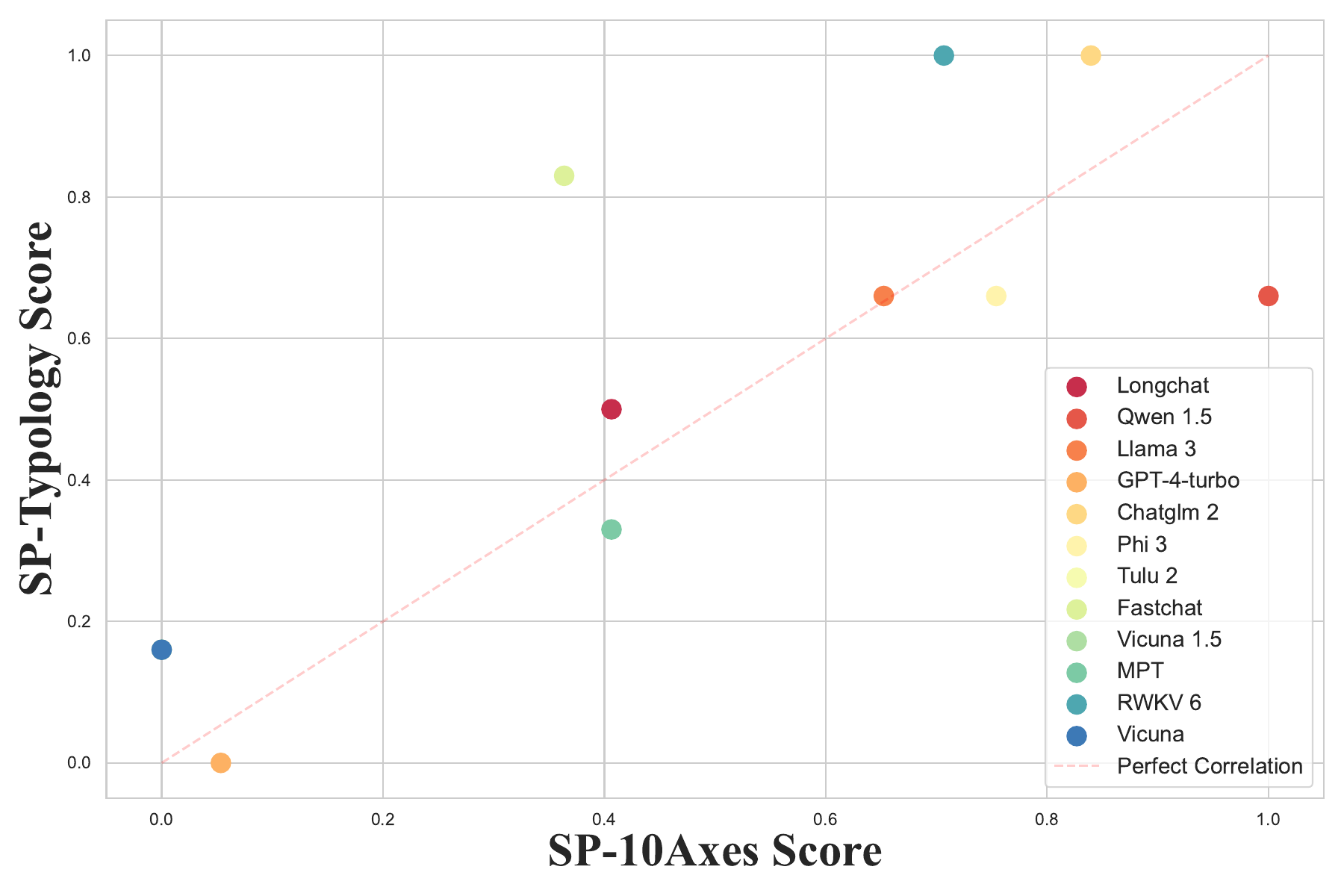}
  \caption{Correlation between the SP-Typology and SP-10Axes}
\label{fig:corr}
\end{figure}

\label{appx:correlation analysis}
Although we rigorously assess the spiritual values of LLMs using similar formats, due to the different institutional origins of these assessments, their objectives are inherently distinct. Specifically, the SP-Typology aims to measure the degree of religiosity in assessments' takers, while the SP-10Axes seeks to categorize individuals into various religious groups. Consequently, it is challenging to assert that LLMs exhibit aligned spiritual values based solely on a direct comparison of the assessment results.

To effectively bridge the two assessments, we focus on the tenth axis of the SP-10Axes: Atheistic vs. Religious. This axis directly corresponds with the objective of the SP-Typology. By mapping and normalizing the LLMs' performance on this axis and the SP-Typology categorization using scores ranging from 0 to 1, we can evaluate whether the LLMs exhibit consistent spiritual values across the two assessments.
The result is shown in Figure~\ref{fig:corr}, which illustrates a moderate alignment between the results of the SP-Typology and the SP-10Axes. The data points, representing different language models, are mostly clustered around the line of perfect correlation, indicating a consistent relationship between the two assessments. This alignment demonstrates the robustness of our evaluation methods' ability in assessing the spiritual orientations of language models.

\section{Quantifying the Spiritual Bias Effect on Social-Oriented Fairness Tasks}
\label{sec:downstream}
Given the premise that LLMs show various spiritual values, it is imperative to investigate how these differences influence impact their downstream performance on social-oriented fairness tasks. 


\subsection{Hate Speech Targeting Different Religions}
We examines the influence on religion-targeted hate speech detection tasks. The \textit{hate-speech-identity} dataset~\cite{yoder2022hate} (21,053 discourses) is adopted and we narrow the scope down into six major spiritual-oriented groups: ``Judaism, Christianity, Islam, Hinduism, Buddhism, and Non-religious''. Details of the mapping method from the dataset's original identity groups to our spiritual-oriented groups are presented in Appendix \ref{appx:hate detection mapping}.
Our evaluation uses F1 score to assess the LLMs' performance in hate detection with regards to each spiritual group. For each evaluated instruction-tuned LLM, we configured the system prompt to position the LLM as an expert in detecting hate speech content. The output was binary, requiring the LLM to respond with either "Yes" or "No," for whether the input text contain hate speech towards a specific spiritual group. 
For our experiment, we choose one representative LLM from each religious group defined in Figure \ref{fig:rel_typology}. The result is displayed in Table \ref{tab:LLM_hate}, which shows an inclination of performance in hate detection task if the LLM falls into a higher spiritual group.

\begin{table*}[ht!]
\centering
\caption{F1-score (in percentage) of selected LLMs on religion-hate speech detection.}
\vspace{-0.3cm}
\begin{tabular}{c|c|cccccc|c}
\toprule
LLM & Type&Buddhism	& Christian &	Hinduism	& Islam	&Judaism	& Non-religious  & Overall \\
\midrule
RWKV6 & Sunday Stalwarts&58.60 & 55.94 & 59.12 & 70.64 & 73.88 & 62.08 & 71.06\\
Llama3 & God Believers & 53.46 & 56.86 & 54.33 & 74.51 & 76.50 & 57.80 & 74.76\\
Qwen2.5 & Diversely Devout&58.49 & 59.49 & 59.45 & 73.67 & 75.99 & 63.50 & 74.23\\
LongChat1.5 & Relaxed Religious&  42.71 & 48.21 & 44.35 & 72.14 & 74.18 & 47.84 & 72.05 \\
Vicuna &Religion Resisters& 20.20 & 12.66 & 17.49 & 7.04 & 14.22 & 21.30 & 9.19\\
Tulu2 &Religion Resisters& 62.33 & 41.14 & 64.14 & 51.81 & 63.43 & 65.65 & 55.67\\
\bottomrule
\end{tabular}
\label{tab:LLM_hate}
\end{table*}

\begin{table*}[ht!]
\centering
\caption{F1-score (in percentage) on religion-hate speech detection before and after further pre-training.}
\vspace{-0.3cm}
\begin{tabular}{cc|cccccc|c}
\toprule
PLM & Train Corpus & Buddhism	& Christian &	Hinduism	& Islam	&Judaism	& Non-religious  & Overall \\
\midrule
GPT2 & N/A & 34.43 & 36.87&	36.30	& 53.28 &	54.54 &	37.92 &	53.30\\
\hline
GPT2 & Bibel & 38.86 &	43.70&	40.96	&61.83	&64.35	&43.42	& 62.36\\
GPT2 & Quran &36.09 &	38.27	&37.54&	56.30&	58.29&	40.77	&56.85\\
GPT2 & Pali &36.42	&38.00	&37.74	&54.19	&55.13&39.37&	54.44\\
GPT2 & Veda & 41.68&	45.41&	43.09&	66.60	&69.50&	46.30&	67.14\\
GPT2 & Tanakh &36.34&	46.17	&36.91&	59.95	&60.23	&44.00	&59.86\\
\bottomrule
\end{tabular}
\label{tab:PLM_finetune_hate}
\end{table*}

\subsection{Effects from Different training sources}
Given the observations in Table~\ref{tab:LLM_hate}, we want to study the effect of fine-tuning with religious literature may have on hate speech detection. We compare the result of a GPT-2-Medium model before and after fine-tuned by Religious Canons of interested religions. The evaluation is a two-step process. In the first step, to elicit GPT-2's ability to response accordingly to hate speech detection task, we employ in-context-learning with four examples from the training data. In the second step, we classify GPT-2's responses using a zero-shot classification pipeline with the facebook/bart-large-mnli model, mapping the generated text to the category: [hate\_speech, no\_hate\_speech, unclear].
The result of the experiment is displayed in table \ref{tab:PLM_finetune_hate}. The findings indicate that, after fine-tuning on religious canons, the performance of LLMs improves on the hate detection task across all canons. Notably, models fine-tuned on the Vedas exhibit the highest improvement, with an increase of approximately 14 percent. This observation aligns with the conclusion from Section 4.1, reinforcing the notion that the spiritual content within religious texts can positively influence the ability of LLMs to perform tasks related to humanistic concerns.
\section{Conclusion}
We have demonstrated that, despite being primarily trained on general domain text, LLMs exhibit varying degrees of spiritual values through comprehensive assessments. Our findings challenge the hypothesis that LLMs are inherently secular. Instead, our experiments show that further exposing LLMs with religious texts can mitigate existing biases and improve performance on tasks related to humanistic concerns, such as detecting hate speech targeted at specific religious groups. These results underscore the importance of careful consideration when developing LLMs in contexts involving spiritual or moral values.

\clearpage
\section*{Limitations}

While our study emphasizes the evaluation of spiritual values and biases in LLMs, we acknowledge several limitations in our approach. One key limitation is the nature of the assessment tools used, which may include behavior-based questions such as "How often do you attend religious services?" Since LLMs lack physical bodies and personal experiences, they can only respond to such questions in a hypothetical manner, which may limit the accuracy of their spiritual self-representation. Future work should consider developing assessment tools tailored specifically to the capabilities and limitations of LLMs in order to better capture their underlying biases in relation to spiritual and moral values.
Additionally, LLMs' responses may be shaped by the in-context learning examples and language used in the prompt, which could inadvertently influence their answers. In the future, auto-prompt generation techniques that minimize prompt bias can be used to ensure more consistent and objective responses.

\noindent \textbf{Disclaims}: We have used AI assistant (\textit{e.g.} chatGPT) to polish manuscript writing.

\section*{Acknowledgment}
Research reported in this work was partially supported by the Atlanta Interdisciplinary Artificial Intelligence Network's 2024 Seed Grant. The content is solely the responsibility of the authors and does not necessarily represent the official views of the the Atlanta Interdisciplinary Artificial Intelligence Network.

\bibliographystyle{ACM-Reference-Format}
\bibliography{software}

\clearpage
\appendix
\appendixpage
\section{List of Evaluated LLMs}
\label{appx:list of llms}

This section provides a detailed list of the LLMs that are evaluated in our study. They are:\\
(a) \textit{Commercial LLMs}:  \\
\indent (a.1) \textbf{GPT-4-turbo}~\cite{openai2024gpt4technicalreport}: a Transformer-based model designed by OpenAI, known for its conversational abilities and fine-tuning with RLHF.  \\
(b) \textit{Open-source LLMs}:  \\
\indent (b.1) \textbf{RWKV-6}~\cite{peng2023rwkvreinventingrnnstransformer}: a unique model that utilizes a Recurrent Neural Network (RNN) architecture, offering efficiency while maintaining capabilities similar to Transformer-based models.  \\
\indent (b.2) \textbf{LLAMA-2}~\cite{touvron2023llama}: high-performing, Transformer-based models developed by Meta, notable for its efficient scaling and fine-tuning for a wide variety of tasks.  \\
\indent (b.3) \textbf{LLAMA-3}~\cite{dubey2024llama3herdmodels}: an improved version of LLAMA-2 by large traiing data, enhanced model architecture, improved hanlding of ambiguity and uncertainty, and enhanced safety features.  \\
\indent (b.4) \textbf{Qwen-1.5-chat}~\cite{bai2023qwentechnicalreport}: a Transformer-based LLM developed by Alibaba, pretrained for up to 3 trillion tokens of multilingual data with a wide coverage of domains and languages. \\
\indent (b.5) \textbf{Phi-3} \cite{abdin2024phi3technicalreporthighly}: a small yet powerful Transformer-based model developed by Microsoft. \\
\indent (b.6) \textbf{Mistral} \cite{jiang2023mistral7b}: a model that leverages grouped-query attention (GQA) for faster inference and sliding window attention (SWA) to handle sequences of arbitrary length. \\
\indent (b.7) \textbf{Gemma} \cite{gemmateam2024gemmaopenmodelsbased}: a light-weight model developed by Google AI. \\
\indent (b.8) \textbf{Vicuna-v1.5} \cite{chiang2023vicuna}: a model trained by fine-tuning llama-13b on user-shared conversations collected from ShareGPT. \\
\indent (b.9) \textbf{LongChat}: a model adapted from Vicuna to account for longer context length. \\
\indent (b.10) \textbf{FastChat-T5}: an open-source chatbot trained by fine-tuning Flan-t5-xl (3B parameters) on user-shared conversations collected from ShareGPT.  \\
\indent (b.11) \textbf{Tulu-2-dpo}\cite{ivison2023camelschangingclimateenhancing}: a fine-tuned version of Llama 2 that was trained on a mix of publicly available, synthetic and human datasets.   \\
\indent (b.12) \textbf{Mpt-chat}: a chatbot-finetuned decoder-style transformer pretrained from scratch on 1T tokens of English text and code developed by MosaicML. \\
\indent (b.13) \textbf{Chatglm2} \cite{glm2024chatglmfamilylargelanguage}: a Transformer-based model fine-tuned for Chinese-language dialogue and language understanding tasks.   \\


\begin{figure}[htbp!]
\forestset{
  L1/.style={fill=white, draw=black, rotate=90, text width=5cm},
  L2/.style={fill=white, draw=black, text width=3.5cm},
  L3/.style={fill=white, draw=black, text width=3.5cm},
  L4/.style={fill=white, draw=black, text width=2cm},
}
\centering
\resizebox{\linewidth}{!}{%
\begin{forest}
for tree={draw,rounded corners,grow'=0,text width=2cm, text centered,edge+={->}},
forked edges,
[,phantom,
    [SP-Typology, L1, fill=olive!25
      [Highly Religious, L2, fill=pink!75, tier=b,
        [Sunday Stalwarts, L3, fill=pink!50, tier=c,
            [ChatGLM, L4, fill=pink!25, tier=d]
            [RWKV-6, L4, fill=pink!25, tier=d]
            [Gemma-7b, L4, fill=pink!25, tier=d]
        ]
        [God-and-Country Believers, L3, fill=pink!50, tier=c,
            [fastchat, L4, fill=pink!25, tier=d]
        ]
        [Diversely Devout, L3, fill=pink!50, tier=c,
            [Llama3, L4, fill=pink!25, tier=d]
            [QWen, L4, fill=pink!25, tier=d]
            [Phi, L4, fill=pink!25, tier=d]
        ]
      ]
      [Somewhat Religious, L2, fill=lime!75, tier=b,
        [Relaxed Religious, L3, fill=lime!50, tier=c,
            [LongChat, L4, fill=lime!25, tier=d]
        ]
        [Spiritually Awake, L3, fill=lime!50, tier=c,
            [Llama2, L4, fill=lime!25, tier=d]
            [CodeLlama, L4, fill=lime!25, tier=d]
            [MPT, L4, fill=lime!25, tier=d]
            [Mamba-2.8b, L4, fill=lime!25, tier=d]
        ]
      ]
      [Non-Religious, L2, fill=cyan!75, tier=b,
        [Religion Resisters, L3, fill=cyan!50, tier=c,
            [Vicuna, L4, fill=cyan!25, tier=d]
            [Tulu, L4, fill=cyan!25, tier=d]
        ]
        [Solidly Secular, L3, fill=cyan!50, tier=c,
            [GPT-4-turbo, L4, fill=cyan!25, tier=d]
        ]
      ]
    ]
]
\end{forest}
}
\caption{Forest Map of SP-Typology Results.}
\label{fig:rel_typology}
\end{figure}

\section{SP-Typology Evaluation}
\label{appx:result of typology}
The license of SP-Typology~\cite{pew2018religiousTypology} is adopted from a scientific publication. The Pew Research Center defines the survey questionnaire license in their website \url{https://www.pewresearch.org/about/terms-and-conditions/}. We only utilize the survey questionnaire and do not use any survey participant private data.
Figure~\ref{fig:rel_typology} shows detailed experimental results of SP-Typology. Individual LLM is ranked from top to button in the order of high spirituality to low spirituality coarse categories. As can be seen, 6 LLMs are categories into highly religious group, 5 LLMs are with somewhat religious group, and 3 LLMs are with with non-religious group.

\begin{table*}[htbp!]
\centering
\caption{SP-10Axes model scores across all categories.}
\vspace{-0.3cm}
\begin{tabular}{lcccccccccc}
\hline
\textbf{Model Name} & \textbf{Catholic} & \textbf{Protestant} & \textbf{Orthodox} & \textbf{Philosemitic} & \textbf{Islamophilic} & \textbf{Buddhist} & \textbf{Prohindu} & \textbf{Pagan} & \textbf{Satanic} & \textbf{Religious} \\
\hline
LongChat 1.5 & 49.01 & 45.05 & 45.11 & 49.17 & 45.54 & 43.32 & 45.97 & 46.05 & 44.87 & 45.97 \\
Qwen 1.5 & \cellcolor{blue!20}40.84 & \cellcolor{blue!40}20.83 & 48.91 & \cellcolor{blue!15}38.74 & \cellcolor{blue!25}34.60 & 42.89 & \cellcolor{blue!30}32.26 & 44.74 & \cellcolor{red!15}58.16 & \cellcolor{blue!25}34.22 \\
ChatGLM 2 & \cellcolor{blue!35}27.23 & \cellcolor{blue!35}29.95 & \cellcolor{blue!35}27.17 & \cellcolor{blue!50}9.50 & \cellcolor{blue!20}36.50 & \cellcolor{blue!35}29.63 & \cellcolor{blue!50}6.05 & \cellcolor{blue!35}26.97 & \cellcolor{blue!35}26.78 & \cellcolor{blue!20}37.39 \\
Phi 3 & \cellcolor{blue!20}38.24 & \cellcolor{blue!30}24.61 & \cellcolor{blue!45}12.36 & \cellcolor{blue!40}18.70 & \cellcolor{blue!35}22.88 & \cellcolor{blue!35}29.63 & \cellcolor{blue!40}14.11 & 50.00 & \cellcolor{blue!30}31.17 & \cellcolor{blue!20}39.09 \\
Tulu 2 & 50.99 & 54.95 & 54.89 & 46.69 & 53.35 & 53.45 & 49.80 & 51.97 & \cellcolor{red!20}58.47 & 54.03 \\
GPT4 Turbo & 49.13 & 40.62 & 36.14 & \cellcolor{blue!35}26.34 & \cellcolor{blue!35}27.90 & \cellcolor{blue!15}35.85 & \cellcolor{blue!35}26.21 & 47.37 & 51.05 & 52.97 \\
Fastchat t5 & \cellcolor{blue!15}40.72 & 46.09 & 47.28 & \cellcolor{blue!20}42.77 & 47.21 & 45.69 & 43.55 & 48.68 & \cellcolor{red!15}58.05 & 46.82 \\
Vicuna 1.5 & 50.99 & 54.95 & 54.89 & 50.83 & 54.46 & 54.53 & 52.62 & 53.95 & \cellcolor{red!10}55.13 & 54.03 \\
Mistral & \cellcolor{blue!20}39.23 & \cellcolor{blue!20}39.32 & \cellcolor{blue!25}34.78 & \cellcolor{blue!35}28.93 & \cellcolor{blue!40}24.55 & 48.82 & \cellcolor{blue!30}30.85 & \cellcolor{blue!15}40.79 & 51.05 & \cellcolor{blue!15}40.78 \\
Mamba & 50.19 & \cellcolor{red!30}57.45 & \cellcolor{red!30}57.85 & 47.79 & 52.42 & \cellcolor{red!10}52.53 & 49.13 & 51.98 & \cellcolor{red!35}59.46 & \cellcolor{red!10}55.12 \\
MPT & 49.01 & \cellcolor{blue!15}43.75 & 50.54 & 45.04 & 45.54 & \cellcolor{blue!20}41.70 & 45.97 & 46.05 & 44.87 & 45.97 \\
RWKV 6 & 47.28 & 54.95 & 54.89 & \cellcolor{blue!15}38.43 & 54.46 & \cellcolor{red!35}59.91 & 54.03 & 48.03 & \cellcolor{red!10}55.13 & \cellcolor{blue!25}40.03 \\
\hline
\end{tabular}
\label{tab:SP-10Axes full result}
\end{table*}

\section{SP-10Axes Evaluation}
\label{appx:result of 10axes}
We obtain the SP-10Axes~\cite{religiousValueTest} from their GitHub repository \url{https://github.com/bannnedb/Religious-values-test}. The data is under the MIT license, and we are allow to use it in this study.
Table \ref{tab:SP-10Axes full result} shows the detailed experimental result of SP-10Axes. A higher score suggests a inclination of opposition towards corresponding category. For example, a score of 70 on the Religious category suggests more "Secularism" than a score of 40. The table employs a conditional color coding scheme to enhance the visual interpretation of the model scores across various religious categories. Values greater than 60 are progressively shaded in red, with the intensity increasing in 10-point increments, emphasizing higher scores. Conversely, values below 40 are shaded in blue, with deeper blue tones for lower scores, also in 10-point increments.

\section{Hate Speech Detection Dataset Label Mapping}
\label{appx:hate detection mapping}

\begin{table*}[htbp!]
\centering
\caption{Mapping of Religion Categories to Original Categories}
\vspace{-0.3cm}
\label{tab:hate_detection_label_mapping}
\begin{tabular}{ll}
\toprule
\textbf{Religion Category} & \textbf{Original Categories} \\
\midrule
Christianity & christians, priests, mormons, catholic priests, cathlolics, \\
             & christians and groups victimized by hitler, catholics, \\
             & catholic folks, catholic \\
\midrule
Islam & muslim kids, muslim women, islamic folks, muslims, islamic, islamics \\
\midrule
Hinduism & hindu folks, hindus \\
\midrule
Judaism & jews, holocaust survivors, jewish victims, jewish folk, \\
        & holocaust survivers, holocaust victims and black victims of slavery, \\
        & german people/jewish people, holocaust survivors/jews, holocaust, \\
        & black jew, the holocaust \\
\midrule
Buddhism & buddhists \\
\midrule
Non-religious & atheists, nonreligious people \\
\midrule
General & all religions, religious people, non-christians \\
\bottomrule
\end{tabular}
\end{table*}

Table~\ref{tab:hate_detection_label_mapping} shows the mapping of the original datasets' targeting categories to our interested religion categories. After the mapping, we get 7 religion categories serve as the targeted hate religion group.

\end{document}